\documentclass[runningheads]{llncs}

\usepackage{eccv}

\usepackage{eccvabbrv}

\usepackage{graphicx}
\usepackage{booktabs}

\usepackage[accsupp]{axessibility}  

\usepackage[pagebackref,breaklinks,colorlinks,citecolor=eccvblue]{hyperref}

\usepackage{orcidlink}

\usepackage{makecell}
\usepackage{colortbl}
\usepackage{mathrsfs}
\usepackage{amsfonts}
\usepackage{graphics} 
\usepackage[normalem]{ulem}
\usepackage{multirow}
\usepackage{xcolor}
\usepackage{pifont}
\usepackage{booktabs}

\usepackage{cuted}
\usepackage{capt-of}

\makeatletter
\newcommand{\printfnsymbol}[1]{%
        \textsuperscript{\@fnsymbol{#1}}%
}
\makeatother

\makeatletter
\def\@fnsymbol#1{\ensuremath{\ifcase#1\or *\or \dagger\or
   \ddagger\or \mathsection\or \mathparagraph\or \|\or **\or \dagger\dagger
   \or \ddagger\ddagger \else\@ctrerr\fi}}
\makeatother

\newcommand{\new}[1]{#1}

\newcommand{\cc}{\color[rgb]{0,0.6,0.3}\ding{52}}
\newcommand{\xx}{\color[rgb]{0.6,0,0}{\ding{55}}}

\begin{document}

\title{BlinkVision: A Benchmark for Optical Flow, Scene Flow and Point Tracking Estimation using RGB Frames and Events}

\titlerunning{BlinkVision: A Benchmark for Correspondence Estimation}

\author{
Yijin Li\inst{1,2}\thanks{Yijin Li, Yichen Shen, and Zhaoyang Huang contributed equally to this work.} \and
Yichen Shen\inst{1}\printfnsymbol{1} \and
Zhaoyang Huang\inst{2}\printfnsymbol{1} \and
Shuo Chen\inst{1} \and
Weikang Bian\inst{3} \and
Xiaoyu Shi\inst{3} \and
Fu-Yun Wang\inst{3} \and
Keqiang Sun\inst{3} \and
Hujun Bao\inst{1} \and
Zhaopeng Cui\inst{1} \and
Guofeng Zhang\inst{1}\printfnsymbol{2} \and
Hongsheng Li\inst{3,4,5}\thanks{Hongsheng Li and Guofeng Zhang are the corresponding authors.}
}

\authorrunning{Y. Li et al.}

\institute{
State Key Lab of CAD\&CG, Zhejiang University \and Avolution AI \  CUHK MMLab \  Shanghai AI Laboratory \and CPII under InnoHK
}




\maketitle

\begin{abstract}
Recent advances in event-based vision suggest that they complement traditional cameras by providing continuous observation without frame rate limitations and high dynamic range which are well-suited for correspondence tasks such as optical flow and point tracking. However, so far there is still a lack of comprehensive benchmarks for correspondence tasks with both event data and images.
To fill this gap, we propose \textbf{BlinkVision}, a large-scale and diverse benchmark with rich modality and dense annotation of correspondence.
BlinkVision has several appealing properties:
\textbf{1) Rich modalities:}
It encompasses both event data and RGB images.
\textbf{2) Rich annotations:}
It provides dense per-pixel annotations covering optical flow, scene flow, and point tracking.
\textbf{3) Large vocabulary:}
It incorporates 410 daily categories, sharing common classes with widely-used 2D and 3D datasets such as LVIS and ShapeNet.
\textbf{4) Naturalistic:}
It delivers photorealism data and covers a variety of naturalistic factors such as camera shake and deformation.
BlinkVision enables extensive benchmarks on three types of correspondence tasks (i.e., optical flow, point tracking and scene flow estimation) for both image-based methods and event-based methods, leading to new observations,
practices, and insights for future research. The benchmark website is \url{https://zju3dv.github.io/blinkvision/}.
\end{abstract}

\setlength{\parskip}{0.2cm plus4mm minus3mm}

\section{Introduction}

\begin{figure*}[bt]
\centering
\includegraphics[width=1.0\textwidth]{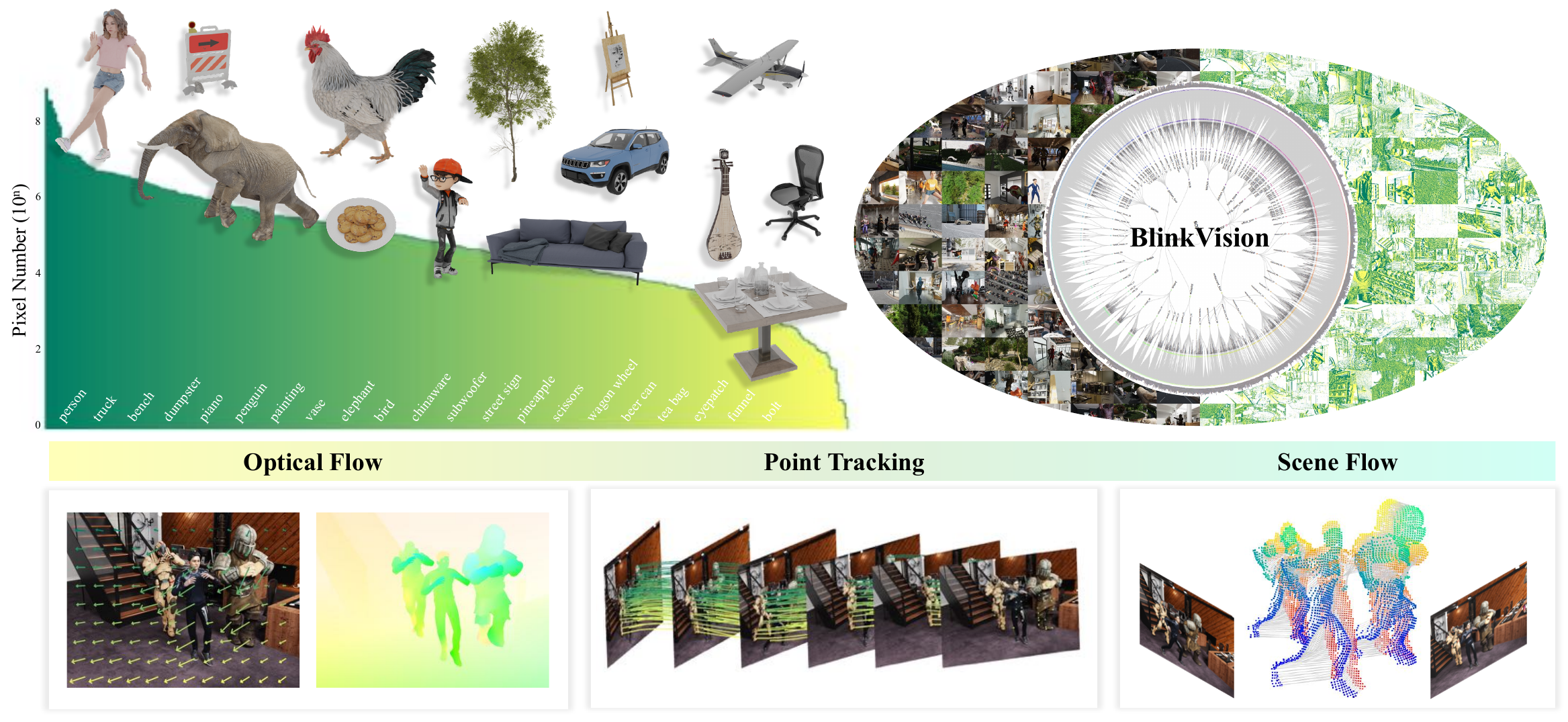}
\captionof{figure}{\textbf{BlinkVision is a large-scale and diverse benchmark with rich modality and dense annotation of correspondence.} It covers 410 daily categories, sharing common classes with popular 2D and 3D datasets. The per-category object distributions, scene structure hierarchy, data samples, and supported applications of BlinkVision are shown in this figure.
\label{fig:teaser}}
\vspace{-1.0em}
\end{figure*}

\begin{table}[tb]
\caption{\textbf{A comparison between BlinkVision and other widely-used benchmarks on pixel correspondence estimation.} ``Occ'', ``Chara'' and ``Cats'' is the abbreviation for occlusion, character and categories, respectively. $\mathrm{R}^{\text{lvis}}$ denotes the ratio of the 1.2k LVIS~\cite{lvis} categories being covered. ``N/A'' denotes the dataset does not provide category labels.}
\scriptsize
\centering
\begin{tabular}{l||ccccccccccc}
\toprule
Datasets & Event & Clean & Final & \makecell{Optical\\Flow} & \makecell{Scene\\Flow} & \makecell{Point\\Tracking} & Occ & Chara & Animal & Cats & R$^{\text{lvis}}$(\%) \\ \midrule
MVSEC~\cite{ev_flownet} & \cc & \xx & \cc & \cc & \xx & \xx & \xx & \xx & \xx &  N/A & 0 \\
DSEC~\cite{dsec} & \cc & \xx & \cc & \cc & \xx & \xx &\xx & \xx & \xx &   N/A & 0  \\
BlinkFlow~\cite{blinkflow} & \cc & \cc & \xx & \cc & \xx & \xx & \cc & \cc & \xx  & 55 & 4.1 \\
EKubric~\cite{rpeflow} & \cc & \cc & \xx & \cc & \cc & \cc & \cc & \xx & \xx & 17  & 0.9 \\
Sintel~\cite{sintel} & \xx & \cc & \xx & \cc & \xx & \xx & \cc & \cc & \cc &   N/A & 0  \\
KITTI~\cite{kitti} & \xx & \xx & \cc & \cc & \cc & \xx & \xx & \xx & \xx &   N/A & 0  \\
FlyingThings~\cite{flyingthings} & \xx & \cc & \xx & \cc & \cc & \xx & \cc & \xx & \xx & 55 & 4.1 \\
Kubric~\cite{kubric} & \xx & \cc & \xx & \cc & \cc & \cc & \cc & \xx & \xx &  17 & 0.9 \\
TAP-Vid~\cite{tap_vid} & \xx & \cc & \xx & \xx & \xx & \cc & \xx & \cc & \xx &   N/A & 0  \\
PointOdyssey~\cite{pointodyssey} & \xx & \cc & \xx & \xx & \xx & \cc & \cc & \cc & \cc &   N/A & 0  \\
BlinkVision (Ours) & \cc & \cc & \cc & \cc & \cc & \cc & \cc & \cc & \cc & 410  & 33.4 \\ \bottomrule
\end{tabular}
\label{tab:dataset_comp}
\end{table}

Modern image-based computer vision technology still cannot match the accuracy and robustness of human vision in many areas.
One possible reason is that traditional cameras suffer from motion blur and limited frame rates, and they often rely on well-lighted conditions. In contrast, event cameras~\cite{event_survey,li2021graph} detect changes in intensity at each pixel as a stream of asynchronous events, which eliminates frame rate limitations and enables operation within a high dynamic range. However, they can not capture fine-grained details as traditional cameras do. Looking at how human vision works, we find that two cells in the human retina, i.e., cones and rods, work similarly to these two types of cameras, respectively. According to the duplex theory of vision~\cite{duplex_theory}, which posits that rods and cones serve different functions, their combination ensures the robustness of human visual processing. Therefore, we believe that combining the advantages of traditional cameras and event cameras can significantly enhance computer vision systems, providing more comprehensive and adaptive vision capabilities.

However, there are currently only a few benchmarks~\cite{dsec,blinkflow,tum_vie} providing both event data and RGB frames, which hinder the development of algorithms that fully exploit the event data and fuse information from both modalities.
This shortfall is particularly prominent in the domain of pixel correspondence estimation~\cite{huang2023flowformer,bian2023context,huang2021vs,yang2023hybrid3d}, i.e., optical flow, scene flow, and point tracking estimation.
Previous benchmarks built for pixel correspondence are either highly biased to specific scenes~\cite{dsec,mvsec} or rather simple~\cite{DVSFLOW16}.
A primary factor is that obtaining such precise pixel-wise annotations for these tasks is expensive.

To boost the research in this area, we present BlinkVision, a synthetic benchmark for optical flow, scene flow and point tracking estimation using RGB frames and events. Our dataset has several appealing properties:
\textbf{1) Rich modalities:}
BlinkVision encompasses three visual modalities: final RGB images, clean RGB images and event data. 
The final RGB images reflect real-world challenges like motion blur and limited dynamic range while the clean RGB images are devoid of such imperfections. The clean RGB images can be seen as the latent images~\cite{pan2019bringing} of event cameras.
\textbf{2) Rich annotations:}
It provides annotations covering optical flow, scene flow, and point tracking. Unlike benchmarks~\cite{dsec,kitti,pointodyssey} which only contain sparse annotations, our data provide dense per-pixel annotations of each image, covering objects including moving cars, deformable characters, and animals.
\textbf{3) Large vocabulary:}
It incorporates 410 daily categories, sharing common classes with widely-used 2D and 3D datasets such as ImageNet~\cite{imagenet}, LVIS~\cite{lvis}, and ShapeNet~\cite{shapenet}, as depicted in Fig.~\ref{fig:teaser}. To our knowledge, our data have the widest variety of objects in existing pixel correspondence benchmarks. This expansive vocabulary is pivotal, enabling rigorous exploration of algorithmic generalization across varied objects.
\textbf{4) Naturalistic:}
BlinkVision employs high-quality assets and rendering tools and thus is able to deliver photorealism data. Besides, it covers a variety of naturalistic factors such as camera shake and deformation.

\new{
To make full use of BlinkVision, we set up a public benchmark website that allows uploading results and provides a public leaderboard.
Besides, we evaluated both existing image-based and event-based methods on three typical correspondence tasks (i.e. optical flow, point tracking, and scene flow estimation). The results reveal new observations and challenges and serve as the baseline for future approaches.
}
Specifically, we first study the robustness of existing image-based methods under large frame intervals and extreme illumination and point out the new challenge for these methods.
Second, the benchmark results on existing event-based methods show that current methods do not fully unleash the potential of event cameras.
Third, we show that fine-tuning existing methods on the training set of BlinkVision significantly boosts the generalizability, demonstrating the vast diversity of BlinkVision.
Finally, the broad diversity and accessible category labels in BlinkVision allow us, for the first time, to analyze the performance of correspondence tasks on different categories.
We believe BlinkVision has the potential to serve as a cornerstone benchmark for advancing the development of more robust computer vision systems.

\begin{figure*}[bt]
\centering
  \includegraphics[width=1\textwidth]{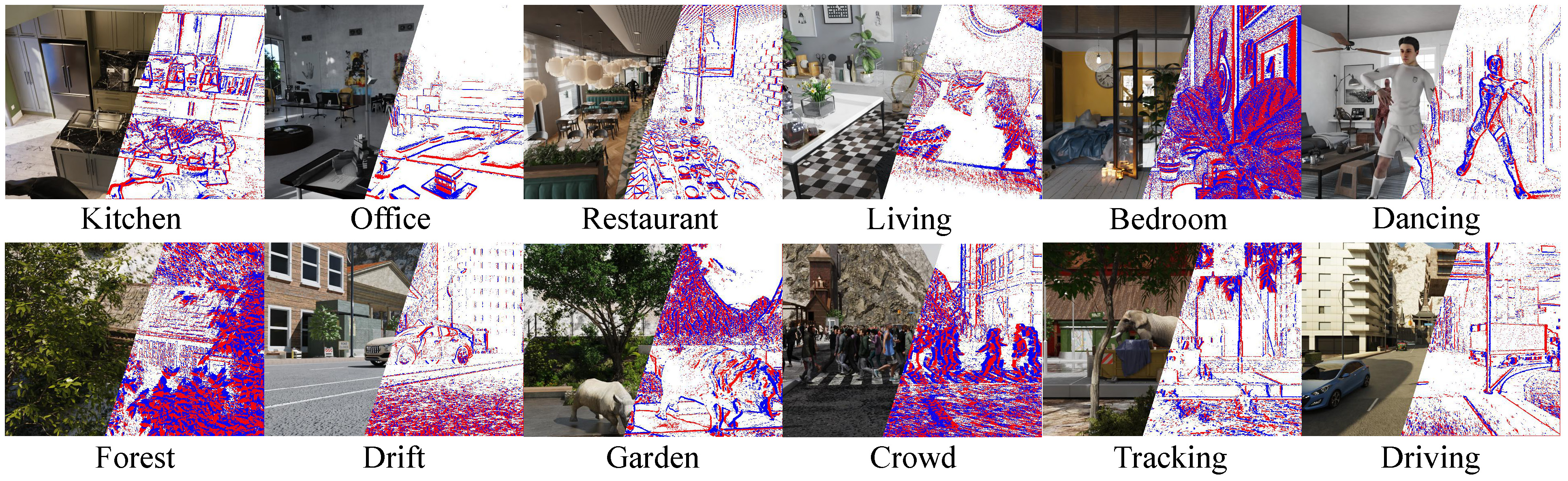}
  \vspace{-1.0em}
  \caption{\textbf{Scenes samples in the proposed BlinkVision benchmark.}}
  \vspace{-1.0em}
  \label{fig:scene}
\end{figure*}

\section{Related Work}
A comprehensive overview and comparison between these benchmarks and BlinkVision is listed in Table~\ref{tab:dataset_comp}.

\noindent\textbf{Image-based optical flow and scene flow benchmarks.}
Early evaluations depended on synthetic datasets, like the famous "Yosemite" sequence~\cite{barron1994performance}.
MPI Sintel~\cite{sintel} was one of the most representative synthetic benchmarks derived from a short open-source animated 3D movie and it became one of the most popular benchmark datasets.
KITTI~\cite{kitti} was almost the most well-known among real data.
It computed ground truth for static scenes through the data from a 3D laser scanner and the ego-motion data of the car.
In a later version, KITTI extended the ground truth to rigidly moving cars by fitting CAD models of cars.
FlyingThings3D~\cite{flyingthings} was a more recent synthetic dataset.
It contained the ground truth of both optical flow and scene flow.
FlyingThings3D and KITTI were the two most commonly used benchmarks in scene flow evaluation.
However, on both datasets, ground truth about scene flow was not available in occluded regions. Furthermore, on KITTI, ground truth for foreground points was either missing or approximated by a fitted car CAD model.

\noindent\textbf{Image-based point tracking benchmarks.}
The first point-tracking dataset was FlyingThings++~\cite{pips}, which was based on FlyingThings3D.
It was initially designed for training the network. Due to the lack of evaluation benchmarks, it was also used for evaluation in the early stages.
Later, Doersch \etal~\cite{tap_vid} proposed a real data benchmark named TAP-Vid, which was based on two real-world datasets: DAVIS~\cite{davis} and Kinetics~\cite{kinetics}. TAP-Vid relied on manual annotation and could not handle occlusions.
PointOdyssey~\cite{pointodyssey} was a newly proposed benchmark that was based on synthetic data and therefore could provide dense ground truth.
However, it did not guarantee that every pixel had ground truth because it only tracked the mesh vertices of the object.
Furthermore, the naturalism of PointOdyssey was limited to its interior parts.
Its outdoor portion had no realistic layout, just a skybox with randomly dropped objects.

\noindent\textbf{Event-based Benchmarks.}
Early event-based optical flow benchmarks~\cite{DVSFLOW16} limited the camera to only rotational motion and inferred ground truth from the rotational motion of the camera.
Two later benchmarks, DSEC~\cite{dsec} and MVSEC~\cite{ev_flownet} computed the ground truth through LiDAR SLAM.
Similar to KITTI, these two benchmarks limited the ground truth to the static elements of the scene. Besides, they had a rather limited motion pattern.
More recently, Li \etal~\cite{blinkflow} proposed a large-scale diversiform synthetic benchmark named BlinkFlow, which presented more challenging scenes and complex motion patterns.
As for the scene flow benchmark, Wan \etal~\cite{rpeflow} recently converted the existing FlyingThings~\cite{flyingthings} and Kubric~\cite{kubric} datasets to event datasets through video-to-event~\cite{video2event} technology. However, due to the limited frame rates in the original video data, this step inevitably produced artifacts in the event data.
Wan \etal also extended the real-world event dataset DSEC with scene flow ground truth.
As for the event-based point-tracking benchmark, existing methods usually employed Structure-from-Motion~\cite{colmap} (SfM) technology to associate keypoints across multiple frames to obtain correspondence ground truth. Event Camera Dataset~\cite{event_camera_dataset} and EDS dataset~\cite{eds_dataset} were two common benchmarks in this area. Limited by the sparsity and accuracy of SfM, these benchmarks did not support the evaluation of arbitrary point tracking.

\section{BlinkVision Dataset}
In this section, we describe how to build BlinkVision, including the scene setup, data rendering, and generation for the ground truth labels.
The process is based on Blender~\cite{blender} as it provides photorealistic rendering and a flexible data interface that allows us to obtain customized correspondence ground truth.
At the last, we introduce the statistics and distribution of the BlinkVision data.

\subsection{Scene Setting}
\new{
Previous synthetic benchmarks~\cite{sintel, spring} generally rely on open-source movies to avoid heavy scene construction. However, these movies are biased and do not cover diverse enough scenarios.
In order to establish a comprehensive evaluation benchmark, we manually assemble a collection of scenarios that are as rich and diverse as possible.
}
We first look for ready-made scenes that are photorealistic and visually diverse.
Specifically, we purchased 40 indoor scenes and 13 outdoor scenes from Evermotion Archinteriors Collection~\cite{evermotion} that cover common scenes such as living rooms, kitchens, offices, bedrooms, restaurants and gardens.
The original scene is static.
To enhance the realism, we procured 29 scanned and high-fidelity human bodies from ActorCore~\cite{actorcore} and 88 artist-designed animals, together with 104 free characters from Mixamo~\cite{mixamo}. These assets are rigged. We re-targeted them to 
\new{
a wide range of motion models (e.g., more than 100 human motions including dancing, walking, talking, etc.) and placed them in various locations.
}
\new{The human's motions mainly come from motion capture~\cite{actorcore} and the animal's motions are designed by the artist to closely emulate their natural motions.}
Furthermore, we built 50 additional outdoor scenes that cover cases such as a drone flying through the forest and a camera following a human or car from the crowd, which are common but rarely included in existing correspondence benchmarks.
\new{
We set the camera trajectory by referencing the shooting trajectories of real-world videos, including handheld shots, car shots, and drone shots that cover non-uniform motion such as sudden stops and sharp turns.
}
Finally, we obtained 80 indoor sequences and 63 outdoor sequences, where 56 indoor sequences and 47 outdoor sequences are for testing and the others are used for fine-tuning.
Some samples of our data are shown in Fig.~\ref{fig:scene}.
\new{
In the supplementary, we also provide video samples.
}
All the assets used in the testing and fine-tuning are totally disjoint, even including trees.

\subsection{Multi-modality Data Rendering}

We use Blender to render brightness $B$ in linear color space. The following simulations of real-world image capturing and event data are both based on the linear color space. We employ tone mapping and gamma correction on $B$ to obtain commonly used sRGB images.
We call these sRGB images ``RGB (clean)'' because they do not suffer from effects like blurs or overexposed. On the contrary, we call the simulated real-world image capture ``RGB (final)''.

\noindent\textbf{Simulation of Real-world Image Capturing.}
Real-world image capture suffers from motion blur and limited dynamic range. While Blender supports the simulation of the former, the latter needs additional processing.
The latter effect usually leads to overexposed or underexposed.
To simulate it, we follow previous work~\cite{hdr2ldr} and employ the following steps:
(1) Random exposure ratio. We uniformly sample the exposure ratio $w$ in the log2 space within $[-3, 3]$ and multiply it by $B$ to simulate underexposed or overexposed, which gives us the augmented HDR image $H=B\times2^w$.
(2) Dynamic range clipping. 
We clip $H$  according to the formulation $\mathcal{C}(H)=\text{min}(H, 1)$. 
This step leads to information loss for pixels in the overexposed regions.
(3) Non-linear mapping.
To align with how humans see a scene, a camera typically uses a non-linear camera response function (CRF) to modify the contrast of the captured image, which can be formulated by $I_n=\mathcal{F}(I_c)$.
We randomly sample CRFs from an existing dataset~\cite{crf_dataset}.
(4) Quantization.
The pixel values are quantized to 8 bits by $\mathcal{Q}(I_n) = \lfloor 255 \times I_n + 0.5 \rfloor / 255$.
This step causes information loss in underexposed and smooth gradient areas.

\noindent\textbf{Event Data.}
Event cameras work by responding to changes in the logarithmic brightness signal (i.e., $L = \log B$) asynchronously and independently for each pixel~\cite{event_survey}. An event is triggered when the change in brightness (either increase or decrease) since the last event at that pixel reaches a threshold of $\pm T$ (with $T > 0$):
\begin{equation}
    p_{k} (L(u_k, t_k) - L(u_k, t_k - \Delta t)) \geq T, 
\label{eq:event-generation}
\end{equation}
where $\Delta t$ is the time since the last event triggered at pixel location $u_k$ and at time $t_k$. $p_k\in \{-1,1\}$ is the polarity of the brightness change. 
To model the event generation process, we need access to a continuous representation of the visual signal for each pixel.
In practice, it is approximated by rendering images at high frame rates for efficiency~\cite{blinkflow,esim}. Events are then synthesized based on frame-by-frame pixel differences.
More specifically, we adaptively sample frames according to ~\cite{blinkflow,esim} to ensure that the maximum pixel displacement between two sampling timestamps is bounded.
In practice, we employ frame interpolation before simulating the event data to further reduce the pixel displacement.
We use DVS-Voltmeter~\cite{voltmeter} to synthesize events because it can simulate complex noise effects (such as noise effects of temperature and parasitic photocurrent), thus generating realistic events.

\begin{figure*}[bt]
\centering
  \includegraphics[width=1.0\textwidth]{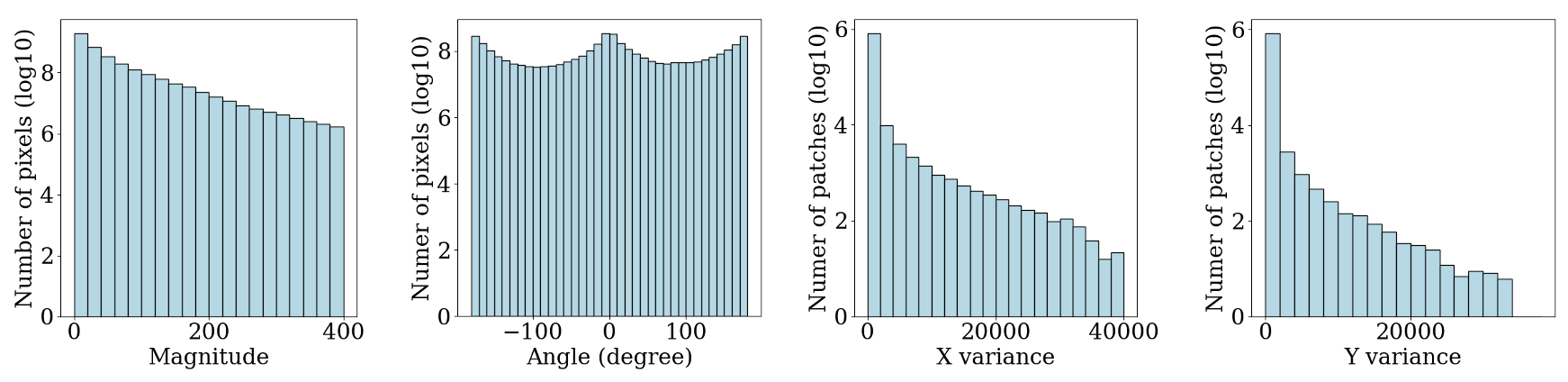}
  \caption{\textbf{Statistics of optical flow in BlinkVision.}}
  \label{fig:flow_stat}  
\end{figure*}

\begin{figure*}[bt]
\centering
  \includegraphics[width=1.0\textwidth]{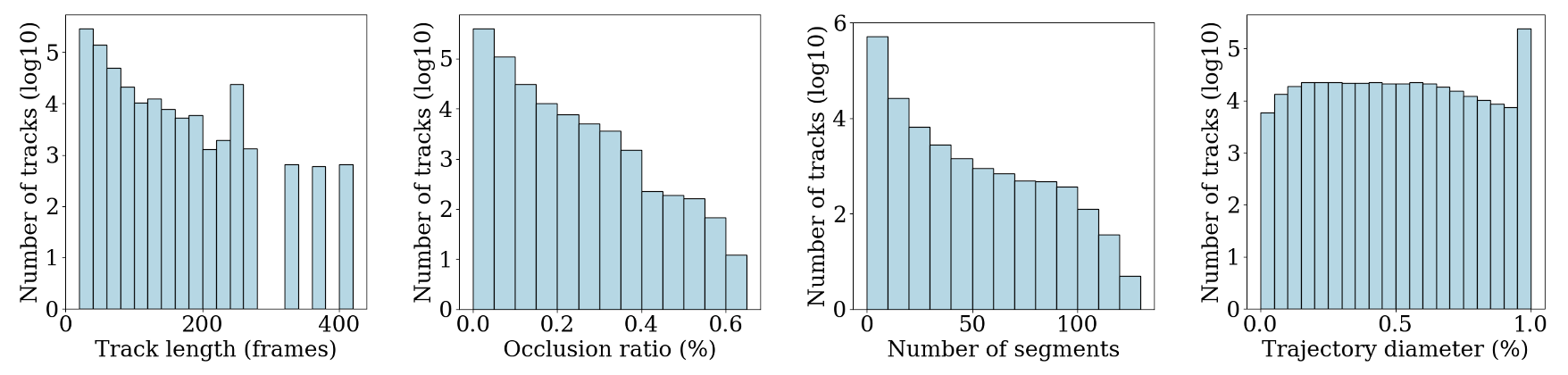}
  \caption{\textbf{Statistics of point trajectories in BlinkVision.}
  To save time, we sample the tracks using a grid with a size of 20. Trajectory segments are defined as contiguous sections of point trajectories, with interruptions caused by occlusion. The diameter is the maximum distance a point moves over time. We clip the trajectory diameter and divide it by the diagonal length of the image to obtain the ratio.
  }
  \label{fig:point_stat}
\end{figure*}

\subsection{Ground Truth Generation}

Blender provides optical flow data between two consecutive frames and a segmentation mask for each object. However, we cannot directly obtain these data from Blender: (1) pixel correspondence across multiple frames; (2) scene flow between frames; (3) semantic category for each object.

\noindent\textbf{Pixel Correspondence.} The data of (1) and (2) can be computed in a unified framework. Given camera poses $\mathtt{T}$ (camera-to-world), depth maps $\mathtt{Z}$ at frames $i$ and $j$, and the camera intrinsic $K$, for the pixel location $\mathtt{u}$ at frame $i$, we first project it to world coordinate: $P=T_i Z_i(\mathtt{u}) K^{-1} \tilde{\mathtt{u}}$, where $\tilde{\mathtt{u}}$ is the homogeneous coordinates of $\mathtt{u}$. Then we re-project the 3D point $P$ to frame $j$ through $d \mathtt{u}_j = K T_j^{-1} P$, where $\mathtt{u}_j$ is the corresponding pixel location and $d$ is the depth value at frame $j$. 
\new{
In this way, we get the pixel correspondence between any two frames and the corresponding depth, i.e., (1) and (2).
}

However, this only works for static objects.
To handle dynamic objects and even those objects with deformation motion such as humans and animals, we bake the face index and barycentric coordinate $(\lambda_1,\lambda_2,\lambda_3)$ of each mesh face into textures.
After rendering the texture, we locate the triangular face that the tracked pixel belongs to (indexed by the rendered face index) and obtain the vertices' 3D position $(V_1,V_2,V_3)$ of that face via Blender's API.
The 3D position of tracked pixels can be computed through $P=\lambda_1 V_1 + \lambda_2 V_2 + \lambda_3 V_3$.
This step replaces the back projection for the static objects.
It is worth noting that we need to triangulate every mesh face in the scenes before baking.

\noindent\textbf{Semantic Labeling.}
BlinkVision contains thousands of objects which makes manual semantic labeling expensive. To this end, we develop an automatic labeling framework that is based on open-vocabulary semantic understanding~\cite{clip,odise}.
First, we pre-define a category list based on LVIS~\cite{lvis}.
The category together with category descriptions is encoded into text
embedding $F_t\in \mathbb{R}^{N\times D}$ with a pre-trained text encoder, i.e., CLIP, where $N$ is the number of categories and $D$ is the feature dimension.
For each object asset, we render it individually to a $224\times 224$ image. The process is rather fast while keeping photorealistic because it does not need to compute complex path tracing between objects. We use the pre-trained image encoder from CLIP\cite{clip} to extract the image embedding $F_i\in \mathbb{R}^{D}$.
The probability of the image belonging to one of the pre-defined categories is computed through $P(c | F_i) = \text{softmax}\left( F_i \cdot (F_t)^{\top} \right)_c$.

\subsection{Dataset Overview}

\new{
BlinkVision consists of 40 training sequences and 103 testing sequences.
The training set provides 107,880 frame pairs for optical flow and scene flow, and 1025 sub-sequences for point tracking.
The frame rate for RGB images and ground truth data is 20 FPS. The frame rate is not practically significant because we adjusted it to ensure sufficient motion between adjacent frames. This adjustment reduces the data size uploaded to the benchmark website.
For each sequence of BlinkVision, we selected the sub-sequences by starting tracking from the first frame. Once the overlap with the reference frame (i.e., how many tracked points remain inside the image) is smaller than 40\%, we start to track a new reference frame. If the overlap is smaller than 20\%, we stop this track.
Additionally, we discard short tracks that are less than 20 frames long because they are less challenging without long-term motion accumulation, which could result in the benchmark easily reaching saturation.
As a result, we generate 1025 sub-sequences where each sub-sequence contains 640$\times$480 dense point trajectories.
For the test set, we selected 12,804 frame pairs for evaluating optical flow and scene flow, and 865 sub-sequences for evaluating point tracking.
BlinkVision includes raw data on depth, normal, camera pose, etc.
At this stage, we focus on pixel-level motion tasks. The additional raw data will be released later, which will benefit other tasks such as depth estimation~\cite{li2022deltar} and SLAM~\cite{liu2023multi,hu2024cp,hu2024cg}.
}

\begin{figure*}[bt]
\centering
  \includegraphics[width=0.9\linewidth]{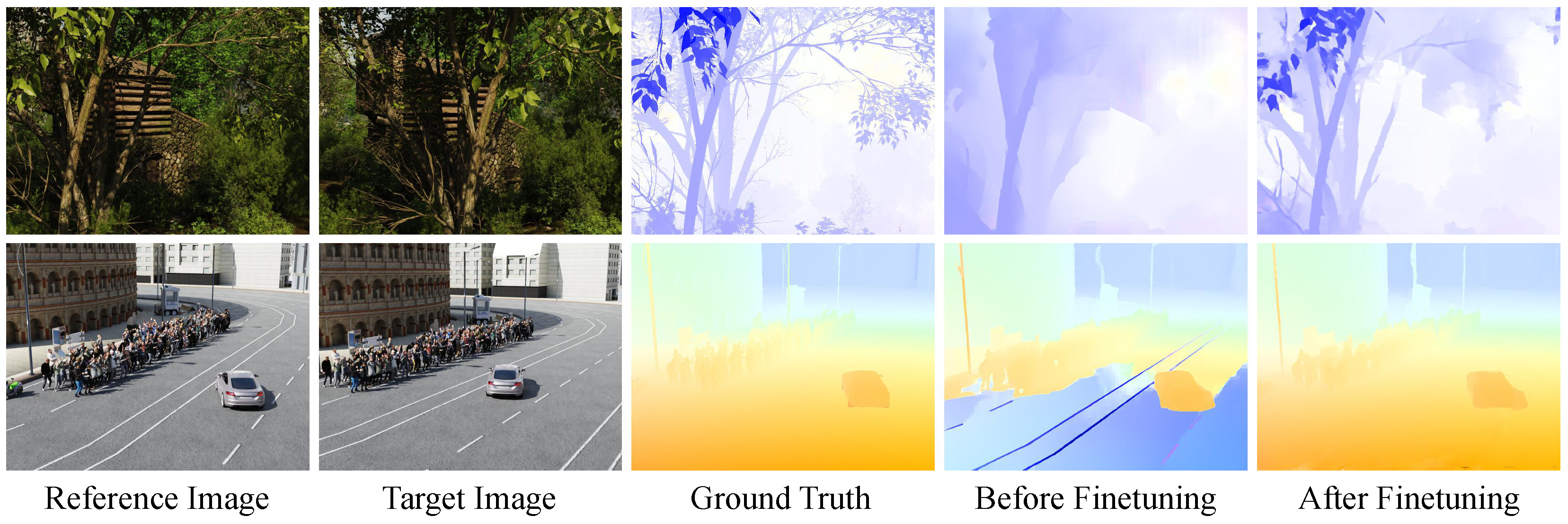}
  \caption{\textbf{Qualitative results of FlowFormer++}~\cite{shi2023flowformer++} before and after fine-tuning on the training set of BlinkVision.}
  \label{fig:flow}
\end{figure*}

Table~\ref{tab:dataset_comp} summarizes key statistics of BlinkVision compared to related works.
BlinkVision contains more than 40K 3D models in 410 categories. It
shares some common categories with 2D and 3D datasets~\cite{lvis,shapenet}
, and bears a huge diversity in semantics, geometry, and appearance, enabling a wide range of research topics.

\new{
We provide statistics of optical flow in Fig.~\ref{fig:flow_stat}.
For benchmarking, we discard those pixels whose flow magnitudes are larger than half of the image diagonal. The latter situation usually occurs when the object is too close to the camera, and its magnitude can even approach infinity. As a result, the magnitude shown in Fig.~\ref{fig:flow_stat} has an upper bound of 400.
The broad distribution in Fig.~\ref{fig:flow_stat} reveals the diverse motions in BlinkVision.
Besides, the diverse distribution of ``X Variance'' and ``Y Variance'' also indicates the presence of complex object shapes in BlinkVision.
We also provide statistics of point trajectories in Fig.~\ref{fig:point_stat}.
BlinkVision allows the evaluation of point tracking with different trajectories' lengths, from small to big, corresponding to different downstream applications like video editing~\cite{neuralmarker,niid_net} or augmented reality~\cite{liu2023multi,ptam}.
The broad distribution of ``occlusion ratio'' and ``number of segments'' indicates that BlinkVision brings diverse and challenging data.
The occlusion ratio has an upper bound that is less than 100 percent due to our strategy of sub-sequences selection.
To quantify the amount of motion a point undergoes, we calculate its trajectory diameter, which is the maximum distance between any two positions of the point over its entire trajectory. We observe that point motion in BlinkVision is diverse and nearly uniformly distributed across the entire image plane.
}

\begin{table}[ht]
\begin{minipage}[c]{0.465\textwidth}
\captionof{table}{\textbf{Optical flow result of RGB-frame-based methods.} ``St.'' denotes Stride. ``\dag'' denotes fine-tuning. ``FF'' denotes FlowFormer~\cite{flowformer} and ``FF++'' denotes FlowFormer++~\cite{shi2023flowformer++}.}
\centering
\scriptsize
\setlength\tabcolsep{2.5pt}
\renewcommand\arraystretch{0.97}
\begin{tabular}{lccccc}
\toprule
\multirow{2}{*}{Method} & \multicolumn{5}{c}{EPE - clean} \\ \cline{2-6} 
 & St. 1 & St. 2 & St. 4 & St. 8 & Avg. \\ \midrule
RAFT~\cite{raft} & 1.83 & 4.62 & 10.18 & 20.46 & 9.27 \\
GMA~\cite{gma} & 1.82 & 4.37 & 9.52 & 18.68 & 8.60 \\
FF~\cite{flowformer} & 1.60 & 3.77 & 7.57 & 16.03 & 7.24 \\
FF++~\cite{shi2023flowformer++} & \textbf{1.54} & \textbf{3.57} & \textbf{7.43} & \textbf{16.26} & \textbf{7.20} \\ \midrule
RAFT\dag~\cite{raft} & 1.33 & 2.96 & 6.39 & 12.72 & 5.85 \\
FF++\dag~\cite{shi2023flowformer++} & \textbf{1.09} & \textbf{2.35} & \textbf{4.88} & \textbf{10.58} & \textbf{4.73} \\ \midrule
\multirow{2}{*}{Method} & \multicolumn{5}{c}{EPE - final} \\ \cline{2-6} 
 & St. 1 & St. 2 & St. 4 & St. 8 & Avg. \\ \midrule
RAFT~\cite{raft} & 2.54 & 5.69 & 11.27 & 22.67 & 10.54 \\
GMA~\cite{gma} & 2.56 & 5.72 & 11.21 & 21.35 & 10.21 \\
FF~\cite{flowformer} & 2.26 & 4.80 & 9.24 & 18.23 & 8.63 \\
FF++~\cite{shi2023flowformer++} & \textbf{2.28} & \textbf{4.83} & \textbf{9.39} & \textbf{19.08} & \textbf{8.90} \\ \midrule
RAFT\dag~\cite{raft} & 1.94 & 3.88 & 7.35 & 15.17 & 7.08 \\
FF++\dag~\cite{shi2023flowformer++} & \textbf{1.66} & \textbf{3.34} & \textbf{6.42} & \textbf{12.73} & \textbf{6.04} \\ \bottomrule
\end{tabular}
\label{tab:rgb_flow}
\end{minipage}
\hspace{0.1cm}
\begin{minipage}[c]{0.523\textwidth}
\captionof{table}{\textbf{Optical flow result of event-based methods.} ``St.'' denotes Stride. ``\dag'' denotes fine-tuning.}
\centering
\scriptsize
\setlength\tabcolsep{1pt}
\renewcommand\arraystretch{1.055}
\begin{tabular}{l||ccccc}
\toprule
\multicolumn{6}{c}{Event} \\ \midrule
Methods & St. 1 & St. 2 & St. 4 & St. 8 & Avg. \\ \midrule
E-RAFT~\cite{eraft} & 2.81 & 7.04 & 17.82 & \textbf{28.60} & 14.07 \\
STE-FlowNet~\cite{ste_flownet} & 2.59 & 5.94 & 13.13 & 41.89 & 15.89 \\
E-FlowFormer~\cite{blinkflow} & \textbf{2.41} & \textbf{5.66} & \textbf{12.53} & 30.45 & \textbf{12.76} \\ \midrule
E-RAFT\dag~\cite{eraft} & 1.68 & 3.63 & 7.48 & 14.20 & 6.75 \\
E-FlowFormer\dag~\cite{blinkflow} & \textbf{1.51} & \textbf{3.00} & \textbf{5.96} & \textbf{13.60} & \textbf{6.02} \\ \midrule
\multicolumn{6}{c}{Event + RGB (clean)} \\ \midrule
Methods & St. 1 & St. 2 & St. 4 & St. 8 & Avg. \\ \midrule
DCEIFlow~\cite{dceiflow} & 3.31 & 14.07 & 34.78 & 61.56 & 28.43 \\
DCEIFlow\dag~\cite{dceiflow} & \textbf{2.19} & \textbf{6.30} & \textbf{12.54} & \textbf{26.11} & \textbf{11.79} \\ \midrule
\multicolumn{6}{c}{Event + RGB (final)} \\ \midrule
Methods & St. 1 & St. 2 & St. 4 & St. 8 & Avg. \\ \midrule
DCEIFlow~\cite{dceiflow} & 4.08 & 14.24 & 34.54 & 61.27 & 28.53 \\
DCEIFlow\dag~\cite{dceiflow} & \textbf{2.52} & \textbf{6.86} & \textbf{13.64} & \textbf{28.45} & \textbf{12.87} \\ \bottomrule
\end{tabular}
\label{tab:event_flow}
\end{minipage}
\end{table}

\section{Benchmark}

\new{
We release all the data except for the ground truth for the test split. For a fair comparison, we create a public benchmark website with leaderboards. We also release the evaluation code for our online benchmarking.
}

\subsection{Optical Flow}

In this section, we comprehensively evaluate the performance of existing optical flow methods on BlinkVision.
Specifically, we first analyze the robustness of image-based methods under large frame intervals and extreme illumination.
Large motion, severe occlusion, and information loss in these extreme cases pose great challenges to existing image-based methods.
In contrast, event cameras naturally depict continuous pixel motion and possess a large dynamic range, thus offering great potential for solving these challenges.
We then benchmark existing event-based methods, including event-only methods and event-RGB fusion methods. We find that existing event-based methods cannot fully unleash the potential of event cameras. We analyze the possible reasons and point to possible opportunities for future research.

\noindent\textbf{Experimental Setup.}
To study the impact of frame intervals on existing methods, we select frames separated by 1, 2, 4, and 8 from the reference frame as target frames.
To reduce the file size of results that are uploaded to the benchmark website, we uniformly sample a reference frame from ten consecutive frames.
As a result, there are 12,804 frame pairs for testing in total.
We follow the previous work~\cite{pats,flowformer,raft} and report the EPE (end-point-error) as the evaluation metric.

\begin{table}
\begin{minipage}[c]{0.58\textwidth}
\begin{minipage}[c]{1\textwidth}
\captionof{table}{\textbf{Point tracking result of RGB-frame-based methods under different frame interval.} ``St.'' denotes Stride.}
\centering
\setlength\tabcolsep{2.1pt}
\renewcommand\arraystretch{0.95}
\scriptsize
\begin{tabular}{l|c||ccccc}
\toprule
\multicolumn{7}{c}{RGB(clean)} \\ \midrule
Method & Metric & St. 1 & St. 2 & St. 4 & St. 8 & Avg. \\ \midrule
\multirow{3}{*}{PIPs~\cite{pips}} & $\delta_\text{avg}$ & 37.77 & 39.34 & 36.34 & 25.50 & 34.74 \\
 & Survival$\uparrow$ & 52.28 & \textbf{55.39} & \textbf{54.55} & \textbf{47.58} & \textbf{52.45} \\
 & MTE$\downarrow$ & 50.77 & \textbf{47.81} & \textbf{49.53} & \textbf{62.42} & \textbf{52.63} \\ \midrule
\multirow{3}{*}{\makecell{PIPs++\\~\cite{pointodyssey}}} & $\delta_\text{avg}$ & \textbf{40.02} & 38.25 & 30.24 & 16.84 & 31.34 \\
 & Survival$\uparrow$ & \textbf{53.44} & 53.13 & 47.57 & 39.71 & 48.46 \\
 & MTE$\downarrow$ & \textbf{45.51} & 51.02 & 59.87 & 73.17 & 57.39 \\ \midrule
\multirow{3}{*}{\makecell{Context-\\TAP~\cite{bian2023context}}} & $\delta_\text{avg}$ & 38.75 & \textbf{39.47} & \textbf{36.51} & \textbf{25.99} & \textbf{35.18} \\
 & Survival$\uparrow$ & 49.98 & 53.09 & 52.61 & 46.18 & 50.46 \\
 & MTE$\downarrow$ & 51.46 & 49.61 & 51.83 & 64.39 & 54.32 \\ \midrule
\multicolumn{7}{c}{RGB(final)} \\ \midrule
Method & Metric & St. 1 & St. 2 & St. 4 & St. 8 & Avg. \\ \midrule
\multirow{3}{*}{PIPs~\cite{pips}} & $\delta_\text{avg}$ & 33.02 &	\textbf{34.44} 	&\textbf{31.56} &	21.77 &	30.20  \\
 & Survival$\uparrow$ & 48.35& \textbf{51.56}&	\textbf{50.72} &	\textbf{44.56}	&\textbf{48.80}  \\
 & MTE$\downarrow$ & 53.94 &	\textbf{51.54}&	\textbf{53.65} &	\textbf{66.14} 	&\textbf{56.32} \\ \midrule
\multirow{3}{*}{\makecell{PIPs++\\~\cite{pointodyssey}}} & $\delta_\text{avg}$ &\textbf{35.83} &	34.24 &	27.11 	&15.32 &	28.12  \\
 & Survival$\uparrow$ & \textbf{50.47} &	50.36 	&45.32 	&38.46 &	46.15  \\
 & MTE$\downarrow$ & \textbf{48.37} &	53.74& 	62.34& 	74.59 &	59.76  \\ \midrule
\multirow{3}{*}{\makecell{Context-\\TAP~\cite{bian2023context}}} & $\delta_\text{avg}$ &33.46 &	34.17 &	31.39 &	\textbf{22.00}& 	\textbf{30.25}  \\
 & Survival$\uparrow$ & 45.58 &	48.91 &	48.53 	&43.06 	&46.52  \\
 & MTE$\downarrow$ &55.09 &	53.76& 	56.33 &	68.11 &	58.32  \\ \bottomrule
\end{tabular}
\label{tab:rgb_particle_stride}
\end{minipage}
\end{minipage}
\hspace{0.1cm}
\begin{minipage}[c]{0.41\textwidth}
\begin{minipage}[c]{1\textwidth}
\captionof{table}{\textbf{Point tracking result of event-based methods.} ``*'' denotes that we track ORB~\cite{orb} feature points rather than grid sampled points. ``DET'' denotes Deep-ETracker~\cite{deep_ev_tracker}}
\centering
\scriptsize
\setlength\tabcolsep{2pt}
\renewcommand\arraystretch{1.24}
\begin{tabular}{l||ccc}
\toprule
\multicolumn{4}{c}{Event} \\ \midrule
Method & $\delta_\text{avg}$ $\uparrow$ & Survival$\uparrow$ & MTE$\downarrow$ \\ \midrule
AMH~\cite{amh} & 18.95 & 37.98 & 75.44 \\
HASTE~\cite{haste} & 18.60 & 37.43 & 77.36 \\
AMH*~\cite{amh} & 26.76 & 41.93 & \textbf{62.48} \\
HASTE*~\cite{haste} & \textbf{27.65} & \textbf{42.87} & 62.52 \\ \midrule
\multicolumn{4}{c}{Event + RGB (clean)} \\ \midrule
Method & $\delta_\text{avg}$ $\uparrow$ & Survival$\uparrow$ & MTE$\downarrow$ \\ \midrule
DET~\cite{deep_ev_tracker} & 21.62 & 35.44 & 80.80 \\
DET*~\cite{deep_ev_tracker} & \textbf{33.46} & \textbf{48.82} & \textbf{60.69} \\ \midrule
\multicolumn{4}{c}{Event + RGB (final)} \\ \midrule
Method & $\delta_\text{avg}$ $\uparrow$ & Survival$\uparrow$ & MTE$\downarrow$ \\ \midrule
DET~\cite{deep_ev_tracker} & 19.05 & 33.52 & 85.00 \\
DET*~\cite{deep_ev_tracker} & \textbf{31.78} & \textbf{48.67} & \textbf{58.79} \\ \bottomrule
\end{tabular}
\label{tab:event_particle}
\end{minipage}
\end{minipage}
\end{table}

\noindent\textbf{Results.}
We benchmark several image-based methods in Table~\ref{tab:rgb_flow}.
We make the following observations:
1) The error generally increases linearly with stride size.
This is expected because doubling the stride yields motion that is twice as fast, resulting in an error that is also expected to be twice as high. 
2) Extreme lighting has a heavy effect on existing image-based methods.
In the supplementary materials, we analyze how different exposure factors lead to performance loss.
3) Fine-tuning on the training split of BlinkVision significantly improves the performance. In Fig.~\ref{fig:flow} we show qualitative results of FlowFormer++ before and after fine-tuning on BlinkVision.
After fine-tuning, the flow predictions are more precise and clear, especially near the object boundary.
\new{
As shown in Table~\ref{tab:rgb_flow}, the benchmark is still challenging for methods after fine-tuning.
}

\begin{figure}
\begin{minipage}[c]{0.55\textwidth}
\centering
  \includegraphics[width=1\linewidth]{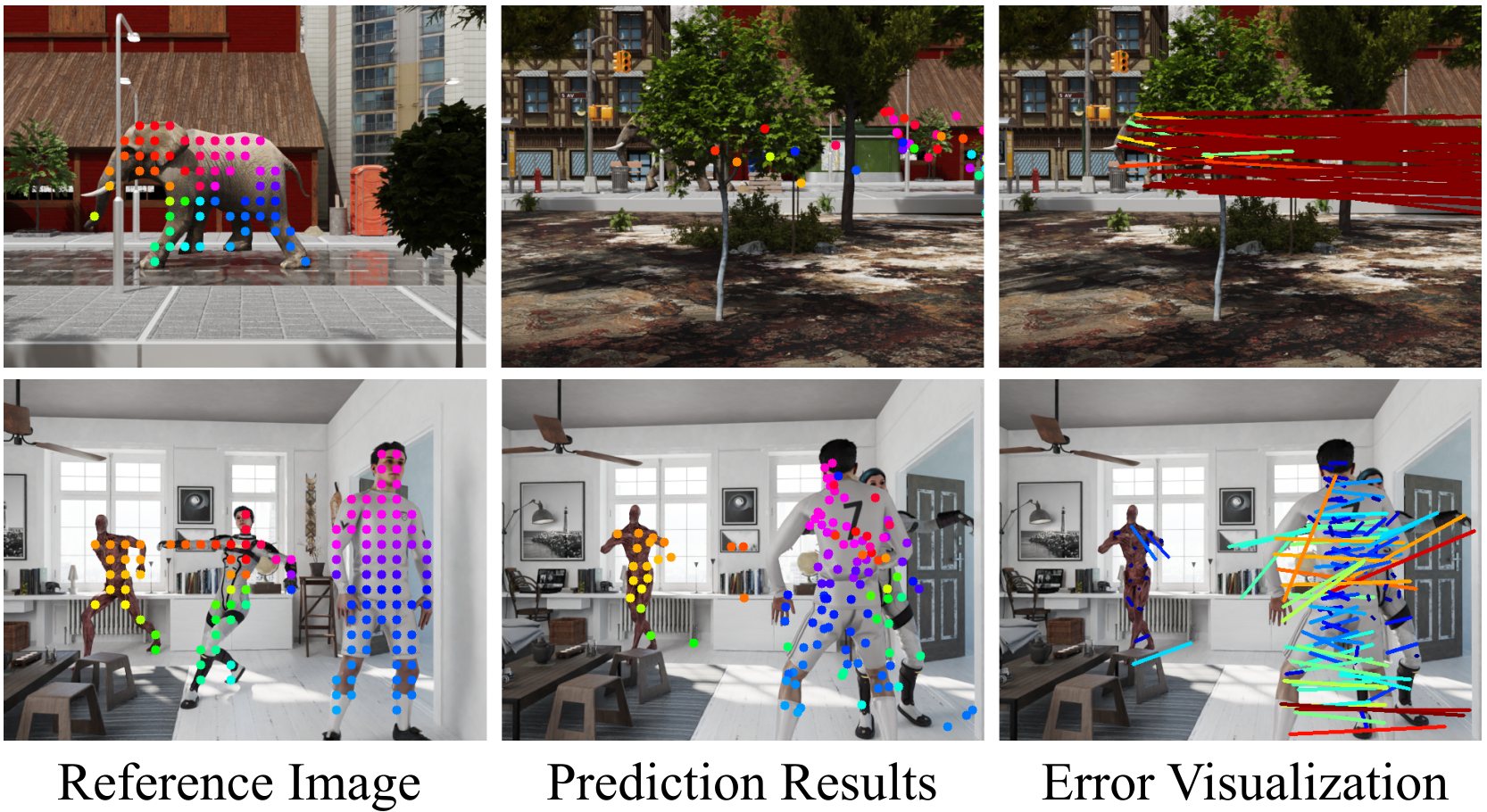}
  \captionof{figure}{\textbf{Qualitative results of PIPs++~\cite{pointodyssey}.} Darker lines in the right figure (e.g., red) indicate larger errors.}
  \label{fig:rgb_particle}
\end{minipage}
\hspace{0.1cm}
\begin{minipage}[c]{0.43\textwidth}
\hspace{-0.3cm}
\centering
  \includegraphics[width=1\linewidth]{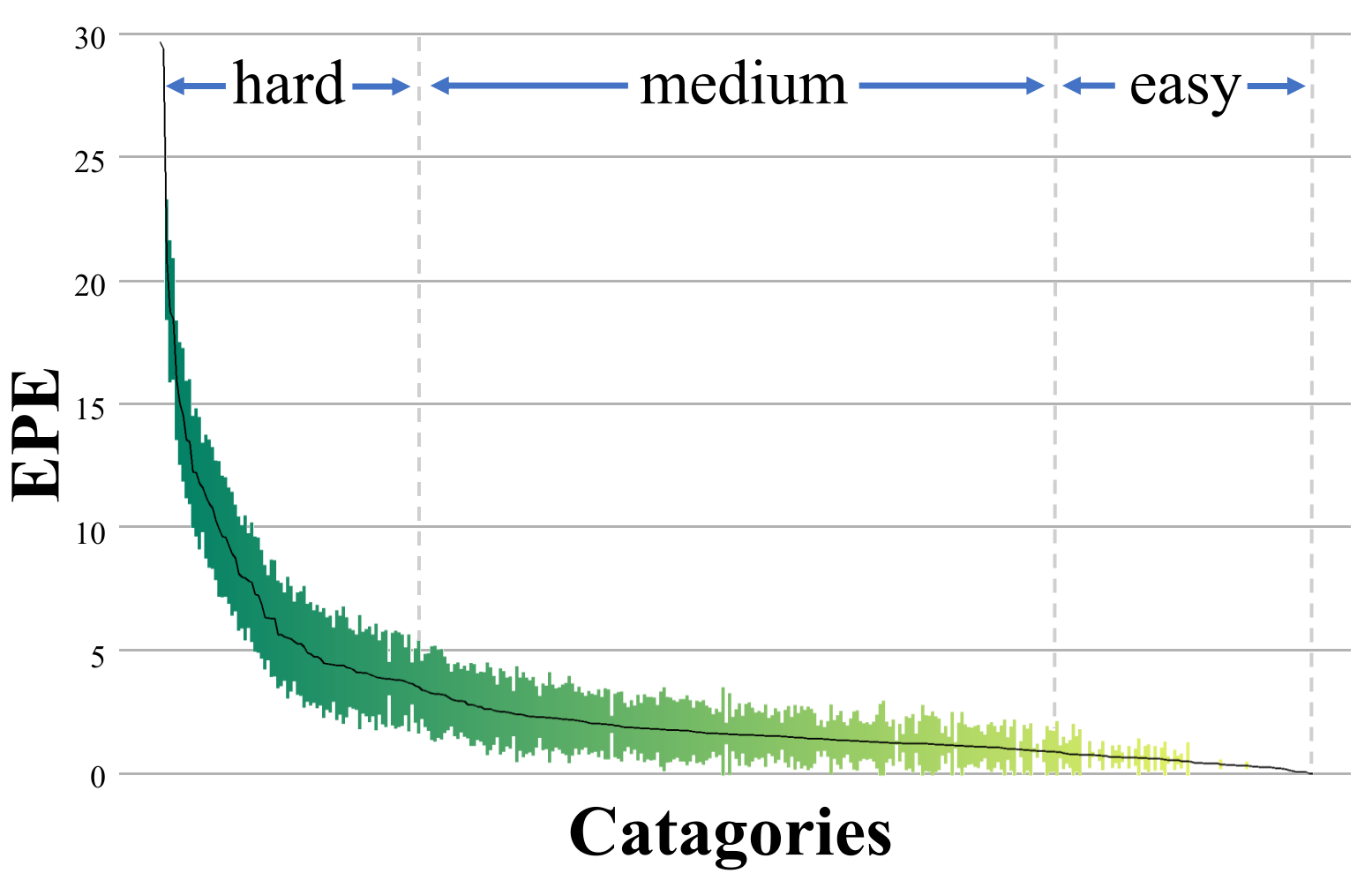}
  \captionof{figure}{\textbf{Performance distribution of FlowFormer++~\cite{shi2023flowformer++}.}}
  \label{fig:semantic}
\end{minipage}
\end{figure}

We report the evaluation of event-based methods in Table~\ref{tab:event_flow}. It is surprising that the performance of existing event-based methods is worse than image-based methods and they also suffer from severe degradation under large frame intervals.
The reasons are mainly twofold.
First, existing training data for events lag behind that of RGB. This can be verified by the performance of E-RAFT~\cite{eraft} and E-FlowFormer~\cite{blinkflow} after fine-tuning. These two methods are far behind RAFT~\cite{raft} before fine-tuning, but exceed the fine-tuned RAFT~\cite{raft} and even FlowFormer++~\cite{shi2023flowformer++} after fine-tuning (in ``Stride-1'', ``Stride-2'' and ``Stride-4'' of the ``final'' case), indicating that the poor performance of the former mainly comes from limited training on existing event-based training data.
Second, existing event-based methods usually convert events to regular 3D voxel grids before processing, which quantizes the event data and leads to information loss. This problem is more serious when dealing with event data within large frame intervals. As a result, it calls for a more powerful representation of event data and new algorithms that can deal with long-range optical flow estimation.

\subsection{Point Tracking}

\begin{table}[t]
\begin{minipage}[c]{0.51\textwidth}
\captionof{table}{\textbf{Scene flow result of RGB-frame-based methods.} ``St.'' denotes Stride.}
\centering
\scriptsize
\setlength\tabcolsep{2.1pt}
\renewcommand\arraystretch{0.75}
\begin{tabular}{l|c||cccc}
\toprule
\multicolumn{6}{c}{EPE-2d} \\ \midrule
Method & RGB & St. 1 & St. 2 & St. 4 & St. 8 \\ \midrule
\multirow{2}{*}{RAFT-3D~\cite{raft3d}} & clean & \textbf{2.15} & \textbf{4.34} & \textbf{8.59} & \textbf{16.08} \\
 & final & \textbf{3.48} & \textbf{7.29} & \textbf{10.74} & \textbf{18.57} \\ \midrule
\multirow{2}{*}{CamLiFlow~\cite{camliflow}} & clean & 2.79 & 6.05 & 12.34 & 23.89 \\
 & final & 4.28 & 8.28 & 16.30 & 30.61 \\ \midrule
\multicolumn{6}{c}{EPE-3d} \\ \midrule
Method & RGB & St. 1 & St. 2 & St. 4 & St. 8 \\ \midrule
\multirow{2}{*}{RAFT-3D~\cite{raft3d}} & clean & 1.91 & 3.83 & 7.25 & 12.18 \\
 & final & 2.02 & 3.96 & 7.39 & 14.21 \\ \midrule
\multirow{2}{*}{CamLiFlow~\cite{camliflow}} & clean & \textbf{1.45} & \textbf{2.76} & \textbf{5.36} & \textbf{9.12} \\
 & final & \textbf{1.53} & \textbf{2.88} & \textbf{5.60} & \textbf{9.81} \\ \bottomrule
\end{tabular}
\label{tab:rgb_scene_flow}
\end{minipage}
\hspace{0.1cm}
\begin{minipage}[c]{0.47\textwidth}
\captionof{table}{\textbf{Scene flow result of event-based methods.} ``St.'' denotes Stride.}
\centering
\scriptsize
\setlength\tabcolsep{2pt}
\renewcommand\arraystretch{1.3}
\begin{tabular}{l|c||cccc}
\toprule
\multicolumn{6}{c}{EPE-2d} \\ \midrule
Method & RGB & St. 1 & St. 2 & St. 4 & St. 8 \\ \midrule
\multirow{2}{*}{RPEFlow~\cite{rpeflow}} & clean & 1.53 & 3.50 & 7.36 & 17.44 \\
 & final & 1.92 & 4.10 & 8.15 & 18.38 \\ \midrule
\multicolumn{6}{c}{EPE-3d} \\ \midrule
Method & RGB & St. 1 & St. 2 & St. 4 & St. 8 \\ \midrule
\multirow{2}{*}{RPEFlow~\cite{rpeflow}} & clean & 5.08 & 9.07 & 15.77 & 25.24 \\
 & final & 5.20 & 9.38 & 15.67 & 25.03 \\ \bottomrule
\end{tabular}
\label{tab:event_scene_flow}
\end{minipage}
\end{table}

In this section, we evaluate the performance of existing methods for long-term point tracking on BlinkVision.
Similar to what we analyzed on optical flow, we analyze the robustness under large frame intervals and extreme illumination, and the gaps between existing image-based methods and event-based methods.

\noindent\textbf{Experimental Setup.}
Although BlinkVision provides per-pixel dense annotations for point tracking, we find that existing methods are not efficient enough to process so much data. As a result, we grid sample the tracked pixels with a grid size of 20 pixels.
We follow PointOdyssey~\cite{pointodyssey} and use $\delta_{\text{avg}}$, median trajectory error (MTE) and survival rate as the evaluation metrics.

\noindent\textbf{Results.}
We benchmark several image-based methods in 
Table~\ref{tab:rgb_particle_stride}.
Similarly, we observe that extreme lighting has a major impact on existing image-based point tracking methods.
In Table~\ref{tab:rgb_particle_stride} we can see that larger frame intervals (such as 4 and 8) severely degrade performance.
However, for PIPs~\cite{pips} and Context TAP~\cite{bian2023context}, increasing the interval from 1 to 2 makes the performance slightly better.
\new{
We guess it might be the limited temporal receptive field of these two methods (only 8 frames) that makes the accumulated error quickly increase when the stride is too small.
}
Some qualitative results are shown in Fig.~\ref{fig:rgb_particle}.
BlinkVision contains many challenging cases that cannot be handled by the SOTA methods.

\begin{table}[t]
\begin{minipage}[c]{0.53\textwidth}
\captionof{table}{\textbf{Cross-dataset evaluation of optical flow methods fine-tuned on BlinkVision.} ``BV'' denotes BlinkVision.}
\centering
\scriptsize
\setlength\tabcolsep{2pt}
\renewcommand\arraystretch{0.9}
\begin{tabular}{l|c||ccc}
\toprule
\multicolumn{5}{c}{RGB} \\ \midrule
\multirow{2}{*}{Method} & \multirow{2}{*}{Data} & Sintel  & Sintel& \multirow{2}{*}{KITTI} \\ 
&  & (clean) &  (final) \\ \midrule
\multirow{2}{*}{RAFT~\cite{raft}} & - & 2.08 & 3.41 & 5.10 \\
 & +BV & \textbf{1.69} & \textbf{3.04} & \textbf{4.66} \\ \midrule
\multicolumn{5}{c}{Events} \\ \midrule
\multirow{2}{*}{Method} & \multirow{2}{*}{Data} & E- & Flying- & E- \\ 
& & Blender&Objects  & Tartan\\ \midrule
\multirow{2}{*}{E-RAFT~\cite{eraft}} & - & 2.66 & 2.9 & 3.27 \\
 & +BV & \textbf{1.92} & \textbf{2.67} & \textbf{2.55} \\ \midrule
\multirow{2}{*}{E-FlowFormer~\cite{blinkflow}} & - & 2.38 & 2.89 & 2.91 \\
 & +BV & \textbf{1.75} & \textbf{2.58} & \textbf{2.36} \\ \midrule
\multicolumn{5}{c}{Events + RGB} \\ \midrule
\multirow{2}{*}{Method} & \multirow{2}{*}{Data} & E- & Flying- & E- \\ 
& & Blender&Objects  & Tartan\\ \midrule
\multirow{2}{*}{DCEIFlow~\cite{dceiflow}} & - & 8.96 & 9.58 & 7.44 \\
 & +BV & \textbf{4.76} & \textbf{2.37} & \textbf{2.88} \\ \bottomrule
\end{tabular}
\label{tab:finetune}
\end{minipage}
\hspace{0.05cm}
\begin{minipage}[c]{0.445\textwidth}
\begin{minipage}[c]{1\textwidth}
\captionof{table}{\textbf{Performance evaluation for typical categories at each difficult level.}``FF'' denotes FlowFormer~\cite{flowformer} and ``FF++'' denotes FlowFormer++~\cite{shi2023flowformer++}. We use the EPE as the evaluation metric.}
\centering
\scriptsize
\setlength\tabcolsep{2.5pt}
\renewcommand\arraystretch{1.35}
\begin{tabular}{l|c||ccc}
\toprule
\multirow{2}{*}{Difficult} & \multirow{2}{*}{Category} & RAFT & FF & FF++ \\ 
 &  & \cite{raft} & \cite{flowformer} & \cite{shi2023flowformer++} \\ \midrule
\multirow{3}{*}{Hard} & person & 9.82 & 8.72 & \textbf{8.61} \\
 & tree & 9.79 & \textbf{9.76} & 10.10\\
 & rhinoceros  & 8.90 & 7.07 & \textbf{6.78} \\ \midrule
\multirow{3}{*}{Medium} & bench & 4.29 & \textbf{3.60} & 3.68 \\
 & globe & 3.43 & 2.52 & \textbf{2.51} \\ 
 & lantern & 2.44 & \textbf{1.92} & 1.95 \\ \midrule
 \multirow{3}{*}{Easy} & notebook & 1.08 & \textbf{0.57} & 0.84 \\
  & coaster & 0.86 & 0.87 & \textbf{0.79} \\ 
 & painting & 0.31 & 0.30 & \textbf{0.26} \\ \bottomrule
\end{tabular}
\label{tab:semantic}
\end{minipage}
\end{minipage}
\end{table}

We show the performance of event-based methods in Table~\ref{tab:event_particle}.
We find that event-based approaches perform particularly poorly.
In addition to the reasons for insufficient training data and model design, we deduce that the receptive field of event-based point-tracking methods is relatively small and therefore cannot perform well on the task of tracking arbitrary points. To verify the claim, we replace the input of grid sampled positions with ORB~\cite{orb} feature points, denoted by ``*'' in the table. The new results perform even better, validating our ideas.
Although the new results cannot be strictly compared with image-based methods, it performs better under extreme frame interval, i.e., ``Stride-8'', which shows the large potential of event-based methods.

\subsection{Scene Flow}

\noindent\textbf{Experimental Setup.}
We use the same pairs as the optical flow benchmark and we follow \cite{raft3d,camliflow} to use 2D EPE and 3D EPE as the evaluation metrics.

\noindent\textbf{Results.}
We benchmark two state-of-the-art image-based methods, i.e., RAFT-3D~\cite{raft3d} and CamliFlow~\cite{camliflow} as shown in Table~\ref{tab:rgb_scene_flow} and show the results of event-based methods in Table~\ref{tab:event_scene_flow}. The conclusions of the optical flow benchmark apply to the scene flow task and present similar challenges for existing methods in this area.
We show qualitative results in the supplementary materials.

\subsection{Cross-dataset Evaluation}

We also fine-tune existing optical flow methods on the BlinkVision training set and then evaluate them on existing representative benchmarks.
The results are shown in Table~\ref{tab:finetune}.
We observe that fine-tuning on BlinkVision brings significant improvement for both image-based methods (2.08 vs. 1.69) and event-based methods (2.38 vs. 1.75 and 9.58 vs. 2.37), which demonstrates the vast diversity of BlinkVision boosts the generalizability of these methods.

\subsection{Performance Distribution on Categories}

Previous methods in optical flow mainly perform evaluations on biased and limited scenarios, which is not comprehensive and robust enough to demonstrate the ability of the methods for different categories of objects in different scenarios.
Thanks to the vast diversity of data and semantic labels provided by BlinkVision, for the first time, we analyze the performance distribution of several image-based optical flow methods on different categories.
The results are shown in Fig.~\ref{fig:semantic}.
The average curve is imbalanced: hard categories usually include complex shapes (e.g., hammocks and shrubs) or with deformable motion (e.g., persons and animals). We thus split the categories into three levels of “difficulty” based on the average curve, and the performance evaluation for typical categories at each level is presented in Table~\ref{tab:semantic}.
We believe such fine-grained analysis helps understand the generalization capacity of the methods.

\section{Conclusion}
We propose BlinkVision, a large-scale diversiform benchmark for three types of correspondence tasks, i.e., optical flow, point tracking, and scene flow estimation using RGB frames and events.
Extensive benchmarks on BlinkVision point to new challenges for existing image-based approaches and show that existing event-based approaches are far from fully unlocking the potential of event cameras.
BlinkVision reveals new observations, challenges, and opportunities for future research into more robust visual systems such as human vision.

\noindent\textbf{Acknowledgment.} This project was funded in part by National Key R\&D Program of China Project 2022ZD0161100, by the Centre for Perceptual and Interactive Intelligence (CPII) Ltd under the Innovation and Technology Commission (ITC)’s InnoHK, by Smart Traffic Fund PSRI/76/2311/PR, by RGC General Research Fund Project 14204021. Hongsheng Li is a PI of CPII under the InnoHK. This work was also partially supported by NSF of China (No. 61932003).

\clearpage

%
%
\bibliographystyle{splncs04}
\bibliography{main}
\end{document}